\documentclass[journal,transmag]{IEEEtran}
\usepackage{graphicx}
\usepackage{amsfonts}
\usepackage{amsmath}
\usepackage{amssymb}
\usepackage{color}
\usepackage{cite}
\usepackage{graphicx,color,overpic}
\usepackage{psfrag}
\usepackage{amssymb}
\usepackage{times}
\usepackage{latexsym}
\usepackage{bm}
\usepackage{cases}
\usepackage{array}
\usepackage{fancyhdr}
\usepackage{setspace}
\usepackage{subfigure}
\usepackage{url}
\usepackage{multirow}
\usepackage{epstopdf}

\usepackage{epsfig}
\usepackage{fancybox}
\usepackage{textcomp}

\DeclareMathOperator*{\argminA}{arg\,min} 

\newcommand{\COS}[0]{\mathrm{cos}}
\newcommand{\x}[0]{\mathbf{x}}
\newcommand{\z}[0]{\mathbf{z}}
\newcommand{\F}[0]{\mathbf{F}}
\newcommand{\f}[0]{\mathbf{f}}

\newcommand{\binning}[0]{\mathbf{b}}

\title{\LARGE \bf
Exploiting Radio Fingerprints for Simultaneous Localization and Mapping
}

\author{Ran Liu, Billy Pik Lik Lau, Khairuldanial Ismail, Achala Chathuranga, Chau Yuen, Simon X. Yang, \\Yong Liang Guan, 
Shiwen Mao, and U-Xuan Tan}

\begin{document}

\maketitle

\pagestyle{headings}
\setcounter{page}{1}
\pagenumbering{arabic}

\begin{abstract}
Simultaneous localization and mapping (SLAM) is paramount for unmanned systems to achieve self-localization and navigation.
It is challenging to perform SLAM in large environments, due to sensor limitations, complexity of the environment, 
and computational resources. We propose a novel approach for localization and mapping of autonomous vehicles using radio fingerprints,
for example WiFi (Wireless Fidelity) or LTE (Long Term Evolution) radio features, 
which are widely available in the existing infrastructure.
In particular, we present two solutions to exploit the radio fingerprints for SLAM.
In the first solution\,--\,namely Radio SLAM, the output is a radio fingerprint map generated using SLAM technique.
In the second solution\,--\,namely Radio+LiDAR SLAM, we use radio fingerprint to assist conventional LiDAR-based SLAM to improve accuracy and speed, while generating the occupancy map.
We demonstrate the effectiveness of our system in three different environments, namely outdoor, indoor building, and semi-indoor environment.
\end{abstract}

\section{Introduction}
\label{Introduction}
SLAM (Simultaneous Localization and Mapping) is essential for unmanned system to carry out high level tasks, such as navigation and exploration in unknown environments \cite{graph_slam_tutorial}.
SLAM allows to build a map and localize an autonomous vehicle at the same time.
With tremendous improvements in computing power and availability of low-cost sensors,
SLAM is now used for many robotics applications.
One example is DARPA subterranean challenge \cite{darpa}, 
which explored innovative technologies to rapidly map, navigate, and search in subterranean domains using robotic systems.

Odometry is commonly used to estimate the position of vehicle.
However, long period of localization often suffers from accumulative error.
SLAM is considered as a fundamental problem in robotics community. 
The core of SLAM is to correct the drifting error of odometry by recognizing if a place has been revisited by the robot.
Efficient SLAM solvers have been introduced and their performance has been evaluated with a variety of sensors 
(e.g., visual camera and LiDAR).
We are now entering an era of robust perception, which considers the long-term operation of the vehicle in large environments.

With the growing popularity of smartphones, 
most existing buildings have been deployed with WiFi and LTE networks for communication purposes \cite{suining_Chameleon}. 
Such radio network can be exploited for localization and mapping with low-hardware requirement and computational cost \cite{BearingSLAM}.
In contrast to visual camera and LiDAR, which measure the similarity of observations by scan matching or feature matching, the radio infrastructure provides an opportunity for SLAM in a cost-efficient way. 
Each base station carries a unique ID, which tells if vehicle is in the proximity.
Particularly, fingerprinting-based approach uses a set of radio signals
to represent location \cite{wifi_crowdsensing} \cite{robust_fingerprint}, which is more accurate when compared to model-based approaches. 

\begin{figure}
\centering
\includegraphics[width=0.49\textwidth]{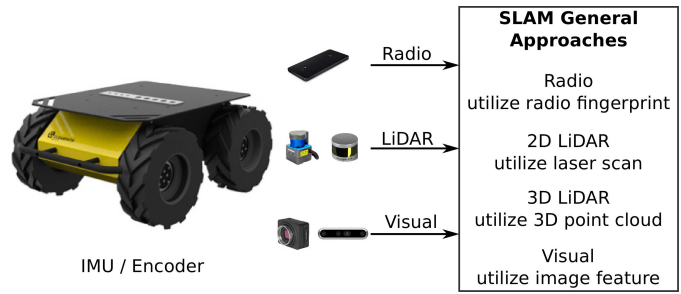}
\caption{
Overview of the SLAM solutions based on different sensors.}
\label{slam_solutions}
\end{figure}

This article provides an overview of the current status of SLAM with emphasis on the challenges and future research directions. 
An illustration of the SLAM approaches based on different sensors is shown in Figure \ref{slam_solutions}.
For Radio SLAM, the SLAM is simultaneous localization and radio
fingerprint mapping, while the LiDAR SLAM is with the mapping of range
scanning. 
Each approach has its advantages as well as challenges. 
For example, LiDAR gives an accurate representation of the environment, but it has issues during loop closure in perceptually-degraded environment. 
Radio fingerprint uses a collection of radio signals to represent location, 
which is shown to be robust again environmental distortions when compared to visual or LiDAR-based approaches, 
but obtaining a precise position estimation is challenging, due to multi-path signal propagation.

To utilize radio features in the existing infrastructure, we present two solutions to use radio fingerprints to perform SLAM. 
In the first solution\,--\,namely Radio SLAM, while generating the trajectory,
we generate a radio fingerprint map using SLAM technique, 
where the produced radio map can be used for localization with the state-of-the-art fingerprinting-based approaches \cite{suining_Chameleon} \cite{ran_iot_2019}. 
In the second solution\,--\,namely Radio+LiDAR SLAM, we use radio fingerprint to assist conventional LiDAR-based SLAM. 
While generating the trajectory, we also generate occupancy map and show that the mapping processing is faster than the conventional approach. 
We conducted experiments in three different environments to illustrate the effectiveness of our approach. 

\section{Sensors and Algorithms in SLAM }
\label{sensors_and_algorithems}
Over the past decades, indoor positioning shows a growing popularity 
due to the increasing demand of location-aware applications \cite{suining_Chameleon}.
SLAM addresses the problem of simultaneous localization and mapping in unknown environment, which has achieved significant achievements over the past years.
In this section, we review the popular sensors and algorithms used for SLAM.

\subsection{Sensors Used for SLAM}
\label{sensors}
Depending on the types of sensors, 
one can classify the SLAM into LiDAR SLAM, visual SLAM, mmWave SLAM, and WiFi SLAM.
LiDAR SLAM uses LiDAR to create structural map of an environment. 
Visual SLAM utilizes camera to construct 3D model of the scene.
Visual SLAM (for example ORB-SLAM3 \cite{OrbSLAM3} and Superpoint \cite{superPoint}) has undergone significant upgrades in recent years due to the evolution of neural networks. 
mmWave carries massive amounts of data at high speed and low latency, which will soon become a fundamental component of 5G-and-beyond communication networks.
The emerging mmWave technology opens new opportunity for localization and SLAM in both wireless and robotics community \cite{mmwave_tutorial} \cite{mmwave_mapping}.
Although SLAM is used for some practical applications, several technical challenges prevent more general-purpose adoption. One challenge is the high computational cost in data processing, particularly when implementing SLAM on compact and low-energy embedded microprocessors.

With the development of Internet of Things (IoT) and wireless communication \cite{distributed_mapping_wifi} \cite{Ismail_case2022}, 
urban environments are deployed with WiFi access points and LTE base stations, which can be exploited for SLAM. 
WiFi SLAM uses wireless signature and odometry for localization and radio mapping in unknown environments.
Ismail \textit{et. al.}\cite{Ismail_case2022}
proposed to estimate the pose of the vehicle with WiFi fingerprint sequence.
Authors in \cite{distributed_mapping_wifi} used WiFi to determine the coarse orientation between LiDAR sub-maps during online distributed mapping.

\subsection{Algorithms in SLAM}
\label{algorithms}
Throughout the years, many techniques and algorithms
have been proposed for SLAM \cite{graph_slam_tutorial}, 
including filtering-based solutions and graph-based solutions.
The choice of the SLAM solution depends on sensors, computational resources, and applications.
Graph-based approaches \cite{graph_slam_tutorial} formulate the SLAM as maximum likelihood estimation and has become one of the favorable approaches. 
The graph-based approach consists of two main modules: front-end and back-end. 
The front-end constructs pose graph based the sensor measurements, while the back-end performs optimization based on the pose graph obtained from the front-end.

When the vehicle operates in the environment for a long time, 
the size of pose graph becomes unbounded, which prevents the optimization of SLAM in real time. 
A common technique is sparsification, which reduces the number of nodes in the graph by pruning \cite{KurzPrunningIROS2021}. 
The wrong loop closures impair the quality of the back-end optimization. 
A number of strategies have been proposed to address this problem.
Pfeifer \textit{et. al.} \cite{Mixture_model} presented max-mixture model for robust sensor fusion by combining expectation-maximization
and non-linear least square optimization.
Several researchers also proposed to remove outliers at the frond-end. 
For example authors in \cite{joint_compatibility} use joint compatibility test to reject spurious constraints.

\section{Challenges and Open research directions}
\label{challenges}
Despite tremendous improvements in SLAM, 
the implementation and deployment of SLAM in practice still face a number of challenges. 
In this section, we discuss the challenges as well as potential research trends in SLAM:

\textbf{Data association with sensor fusion}:
The introduction of novel sensors innovates new applications in SLAM.
The choice of sensors depends on the applications and environments. 
It is difficult to rely on a single sensor to achieve effective front-end sensor fusion due to the sensor limitations.
For example, LiDAR enables the creation of occupancy map of an environment, 
but LiDAR does not work well in perceptually-degraded environments (for example long corridors or tunnels).
In our approach, 
each radio base station has a unique ID. 
This makes the data association easier, as entering the same area can be known by the base station ID.

\textbf{Mapping in large environment}:
With a continuous operation in large scale environment, 
the size of map may become unbounded, which prevents 
the loop closure detection in real time.
The design of efficient SLAM solutions with acceptable memory and computational cost is necessary.
One solution is to develop efficient similarity searching algorithms to find potential loop closures. One example is Faiss library\footnote[1]{https://faiss.ai/} from Facebook, which allows fast similarity search of dense vectors.
Our approach uses radio signature as features to assist the conventional LiDAR SLAM to create a map in large environment.
Another solution is to perform distributed SLAM with multiple robots. 
Authors \cite{Pairwise_outlier_rm} proposed a mechanism called pairwise consistency maximization to filter out spurious loop closures between robots.

\textbf{Failure recovery mechanism}:
Incorrect data association ruins the optimization in SLAM. 
To address this issue, researchers have proposed several techniques in the back-end 
to deal with outliers \cite{Mixture_model} \cite{joint_compatibility}.  
These methods determine the validity of loop closure by checking the residual error during optimization.
Despite the efforts made on the back-end optimization, 
current SLAM solutions are susceptible to false loop closures, which lead to poor estimation 
and prevent the exclusion of spurious loop closures afterwards.
Therefore, a mechanism to recover from such failures is a future research trend. 

\textbf{Semantic mapping}:
In contrast to the geometric representation of the environment, 
semantic mapping \cite{Semantics_mapping} provides an opportunity to understand the scene and interact with the environment.
This technique represents the scene in an unambiguous and informative way,
enabling large scale autonomy and robust perception in highly unstructured environments.
The development of semantic mapping is still at its early phase 
and new tools for example deep learning will obviously hasten the progress of semantic SLAM in real applications.

\begin{figure}
\centering
\includegraphics[width=0.495\textwidth]{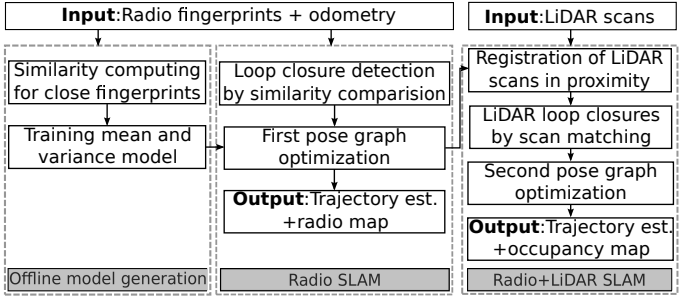}
\caption{Overview of Radio SLAM and Radio+LiDAR SLAM.
}
\label{fig_radio_lidar_slam}
\end{figure}

\section{Radio SLAM and Radio+LiDAR SLAM}
\label{efficient_radio_lidar_slam}
Radio signals are commonly used for localization in many commercial and industrial applications \cite{suining_Chameleon}. 
We propose to use radio fingerprints as alternative to the conventional range-based or visual-based SLAM. 
We present two solutions to use radio fingerprints for SLAM. 
In the first solution\,--\,Radio SLAM, we use radio fingerprint to estimate the trajectory of a robot and generate a radio map of the environment. 
In the second solution\,--\,Radio+LiDAR SLAM, we incorporate LiDAR by scan matching to produce a more accurate trajectory and generate an occupancy map of the environment.
An overview of the proposed approach is shown in Figure \ref{fig_radio_lidar_slam}. 
The similarity check and SLAM are working in parallel in the proposed approach. The similarity check is used to determinate if two locations are in the same place (i.e., loop closures). Meanwhile, the SLAM takes the loop closures as inputs to optimize the trajectory.
Our approach provides an efficient way to perform SLAM in large scale environment, while the conventional LiDAR-based SLAM falls short due to the lack of robust data association algorithms.

\subsection{Overview of Graph-based SLAM}
\label{slam_overview}
Graph-based approach formulates the SLAM as maximum likelihood estimation. 
We denote the pose of robot as $\x=\{x,y,\theta\}$, 
where $x$ and $y$ are the 2D location and $\theta$ is the heading of the robot.
The sensors carried by the robot are used to infer the constraints. 
Graph optimization aims to find the best configuration of nodes to minimize the following equation given constraints $\mathbb{C}$:
\begin{equation}
\begin{split}
\argminA_{\x} \sum_{(\x_i,\x_j) \in \mathbb{C}} & {(\z_{ij}-\hat{\z}_{ij}(\x_{i},\x_{j}))}^\intercal {\Sigma_{ij}}^{-1} \\
&\times(\z_{ij}-\hat{\z}_{ij}(\x_{i},\x_{j})),
\end{split}
\label{eq:optimization}
\end{equation}
where the subscript $i$ and $j$ are used to denote the time reported from the sensor. 
We use $\z_{ij}$ to represent the constraint and its associated weight is denoted as covariance matrix $\Sigma_{ij}$.
The constraint can be either successive odometry or loop closure inferred from non-sequential observations.
Two types of loop closures are involved in optimization, namely radio-based loop closures and LiDAR-based loop closures. 
The former is represented as the distance, which is determined by the fingerprint similarity as shown in Section \ref{radio_slam}.
The latter is denoted as the rigid body transformation by LiDAR scan matching.
We consider a pair $<i,j>$ as WiFi loop closure if the similarity score is over a threshold. We treat a pair $<i,j>$ as LiDAR loop closure if the scan matching satisfies a pre-defined condition. 
Those loop closures (i.e., WiFi loop closures and LiDAR loop closures) and odometry are used as constraints to optimize the trajectory of the robot through pose graph optimization as shown in Equation \ref{eq:optimization}.
\subsection{Fingerprint Similarity}
\label{similarity}
Given a received signal strength (RSS) from a base station, it is straightforward to know if an area has been visited by the robot,
since each reported RSS is associated with a unique ID.
Estimating the precise transformation between two fingerprints turns out to be tricky, 
since radio signal neither reports distance nor bearing information.
In our approach, the closeness of two locations is determined by the fingerprint similarity.
We train a model to represent the distance and variance of radio-based loop closures given the similarity of two location fingerprints, which will be detailed in Section \ref{radio_slam}. 

We represent a fingerprint at pose $\x_t$ as a pair $\F_t={(\f_t,\x_t)}$.
$\f_t$ consists of RSS from $L$ access points (APs) or base stations: $\f_t=\{ f_{t,1},...,f_{t,L} \}$.
Let $L_i$ and $L_j$ denote the number of detections in $\f_i$ and $\f_j$, respectively.
$L_{ij} = \left | \f_i \cap \f_j \right |$ represents the common APs or base stations in $\f_i$ and $\f_j$. 
We use a cosine similarity metric to measure the similarity $\COS(\F_{i},\F_{j})$ between two fingerprints $\F_i$ and $\F_j$:
\begin{equation}
\begin{split}
s_{ij}=\COS(\F_{i},\F_{j})=\frac{\sum_{l=1}^{L_{ij}}{f_i^l f_j^l}} 
{ \sqrt {\sum_{l=1}^{L_i}{(f_i^l)^2}} \sqrt{\sum_{l=1}^{L_j}{(f_j^l)^2}} }
\end{split}
\label{eq:similarity}
\end{equation}

\subsection{Radio SLAM based on Fingerprint Similarity}
\label{radio_slam}
Parameterization of the constraint is required to optimize the pose graph. 
For odometry-based constraint, the parameter is obtained from the motion model. 
We need to derive a model to represent the distance and uncertainty given two fingerprints.
Our solution is to train such model by passing over odometry and radio recordings. 
Although odometry error accumulates over long period, 
it is sufficiently small for a short distance travelled.
Therefore, we compute the degree of similarity for close fingerprint pairs, 
which are annotated with the distance determined by the odometry. 
As a result, we obtain a set of $K$ training samples: $\{s_k, d_k\}_{k=1}^K$, 
where $s_k$ is the fingerprint similarity and $d_k$ is the physical distance of the fingerprint pair.
We then train a model, which features the mean distance $\mu(d|s)$ and variance $var(d|s)$ given a similarity value $s$ by binning:
\begin{equation}
\begin{split}
& \mu(d|s)=\frac{1}{ c (\binning (s,r)) } \sum_{k \in \binning(s,r)} { d_k}^2 \\
& var(d|s)=\frac{1}{ c (\binning (s,r)) } \sum_{k \in \binning(s,r)} { (d_k-\mu(d|s))}^2,
\end{split}
\label{eq:modelling}
\end{equation}
where $\binning (s,r)$ denotes the samples that sit in the interval $r$ around a similarity value $s$.
$c(\cdot)$ counts the number of samples.
Equation \ref{eq:modelling} gives the covariance matrix for radio-based loop closure for Equation \ref{eq:optimization}.
To optimize the trajectory with Equation \ref{eq:optimization}, 
we need to find the constraints for non-consecutive poses. 
We compute the similarity $s_{ij}$ between two recorded fingerprints $F_i$ and $F_j$ 
if the distance travelled by the robot is larger than a pre-defined thresholds (100 meters). 
The distance $z_{ij}$ is obtained based on the pre-trained model in Equation \ref{eq:modelling}.
We add a tuple $<\x_i, \x_j, z_{ij}>$ as a loop closure if the similarity $s_{ij}$ exceeds a threshold $\vartheta_s$.
Based on the radio loop closures, 
we optimize Equation \ref{eq:optimization} using g2o with Levenberg-Marquardt solver \cite{graph_slam_tutorial}.
\begin{table*}[]
\centering
\caption{Description of the data collected in three environments and accuracy evaluation of pure odometry, Wifi SLAM, and LTE SLAM (mean and standard deviation in meters).}
\begin{tabular}{|c|c|c|c|c|c|c|c|}
\hline
Environment  & \begin{tabular}[c]{@{}c@{}}Traj. \\length\,(m)\end{tabular}& \begin{tabular}[c]{@{}c@{}}Traj. \\duration\,(s)\end{tabular} & \begin{tabular}[c]{@{}c@{}}WiFi MAC per\\scan/total\,MAC\end{tabular} & \begin{tabular}[c]{@{}c@{}}LTE CellID per\\scan/total\,LTE\end{tabular} &   \begin{tabular}[c]{@{}c@{}}Odom.\\(m)\end{tabular}      &   \begin{tabular}[c]{@{}c@{}}WiFi SLAM\\(m)\end{tabular}& \begin{tabular}[c]{@{}c@{}}LTE SLAM\\(m)\end{tabular}\\ \hline
SUTD\,outdoor & 5171.3  & 5494.7  & 88.23/4488     & 18.74/170  & 75.06    & 8.82$\pm$3.94    & 18.38$\pm$10.70 \\ \hline
SUTD\,indoor  & 1114.9  & 3092.1  & 118.34/1921    & 19.49/193  & 10.01  & 3.69$\pm$2.21     & 5.18$\pm$2.59 \\ \hline
NTU\,semi-indoor       & 1072.6  & 2583.9  & 48.88/1046     & 16.59/85  & 51.98   & 4.48$\pm$2.98     & 12.22$\pm$10.95 \\ \hline
\end{tabular}
\label{table_evaluation}
\end{table*}
\subsection{Radio+LiDAR SLAM}
\label{radio_lidar_slam}
Based on the trajectory obtained from Radio SLAM, 
we further refine the trajectory by fusing LiDAR scans. 
This is achieved through a second pose graph optimization by considering LiDAR constraints (as shown in Figure \ref{fig_radio_lidar_slam}).
By LiDAR scan registration, a precise transformation between two poses can be computed. 
A common method is Iterative Closest Point (ICP), 
which minimizes the sum of square distance (i.e., fitness score) between correspondences in two scans. 
We consider the following two manners to incorporate LiDAR measurements:
\begin{itemize}
\item \textbf{Registration of LiDAR scans in proximity:} we perform ICP for the consecutive LiDAR scans to correct the short term odometry errors. 
The odometry measurement is used as an initial guess of ICP algorithm to avoid the convergence to local minima.
\item \textbf{LiDAR loop closures:} based on the optimized trajectory from Radio SLAM, 
we perform the LiDAR scan matching if the displacement between two poses is smaller than a threshold. 
We consider a LiDAR loop closure if the fitness score and the matching points satisfy certain criteria. 
This step determines loop closures between non-consecutive poses with a better accuracy when compared to radio fingerprints.
We perform the second pose graph optimization by minimizing Equation \ref{eq:optimization} again but considering the new constraints from LiDAR scan matching. 
The output of Radio+LiDAR SLAM is occupancy map (the same as the conventional LiDAR SLAM), which can be used for the navigation and path planning for autonomous vehicles.
\end{itemize}

\begin{figure}
\centering
\includegraphics[width=0.40\textwidth]{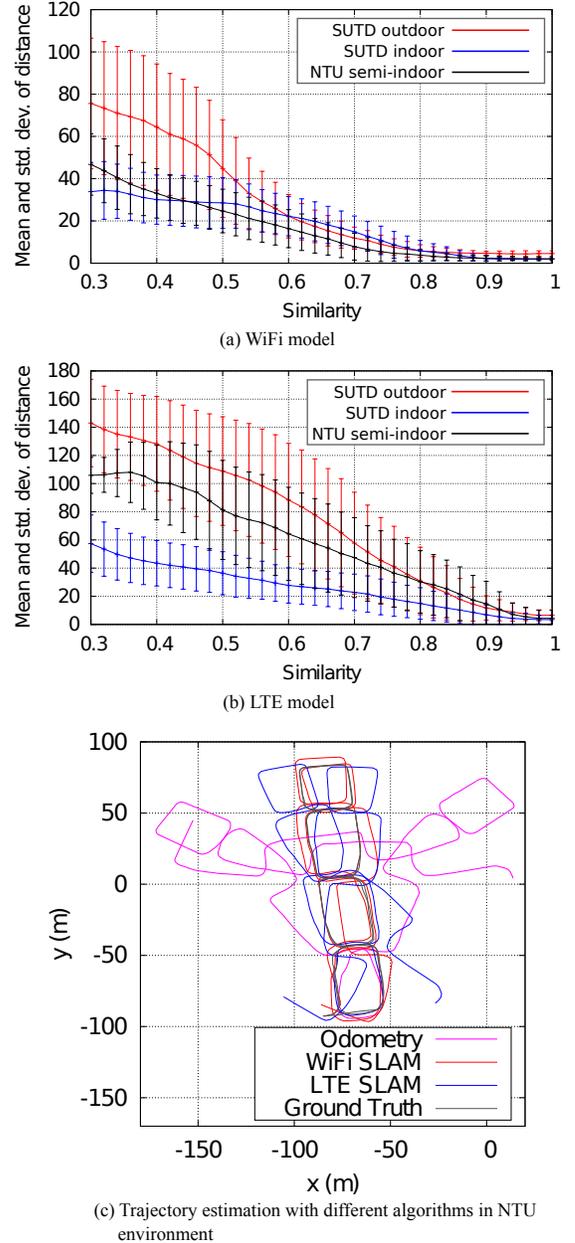}
\caption{A comparison of the models in three different environments 
and visualization of ground truth, odometry, LTE SLAM, and WiFi SLAM in NTU semi-indoor environment.
}
\label{fig_models_track}
\end{figure}

\section{Experimental Results}
\label{experimental_evaluations}
\subsection{Experimental Setup}
\label{experiment_setup}
The experiments were performed using a Clearpath Husky vehicle. 
We evaluated the effectiveness of the proposed algorithm in three different environments, 
which consist of Singapore University of Technology and Design (SUTD) campus (outdoor,  approx. 75000\,m$^2$), 
SUTD building (indoor, approx. 5000\,m$^2$), and Nanyang Technological University (NTU) carpark (semi-indoor, approx. 5100\,m$^2$).
Five Xiaomi Max3 smartphones were placed on the robot to scan WiFi access points and LTE base stations. 
Each phone is configured to scan one LTE band in Singapore. 
Wheel odometry were recorded at a frequency of 10Hz. 
Hokuyo UST-20LX LiDAR was used for LiDAR scans. 
In outdoor experiment, we obtain the ground truth by Android location service.
For indoor environments, 
we have created an indoor map of the environment based on GMapping\footnote[1]{http://wiki.ros.org/gmapping/}. 
With the created indoor map, we apply AMCL (Adaptive Monte Carlo Localization) for ground truth based on probabilistic algorithms.
Particularly, AMCL tracks the 2D pose of the robot against a known map based on particle filtering. The algorithm performs pose prediction and weights update of particles based on odometry and 2D LiDAR, respectively. During our implementation, the map resolution is set to be 0.05m and position of the robot is known during the initialization of the particle filtering.
Throughout the experiments, the robot travelled at an average speed of 0.4m/s. 
Table \ref{table_evaluation} summarizes the data collected in three environments. 

The models of WiFi and LTE generated with a binning size of 0.05 are visualized in Figure \ref{fig_models_track}(a) and Figure \ref{fig_models_track}(b), respectively.
As can be seen from these figures, given a similarity value, LTE model gives large distance and standard deviation 
when compared to WiFi model.
For NTU environment, we obtain mean distance of 30.25m and standard deviation of 22.13m of LTE model given a similarity value of 0.8, 
which are larger when compared to WiFi model (mean distance of 3.79m and standard deviation of 2.70m).
To detect radio-based loop closures, 
we use a similarity threshold of $\vartheta_s$=0.7 for both WiFi and LTE. 

\subsection{Evaluation of Radio SLAM}
\label{experiment_wifi_slam}
The comparison of the track between odometry, LTE SLAM, WiFi SLAM, 
and ground truth in NTU environment are visualized in Figure \ref{fig_models_track}(c). 
Table \ref{table_evaluation} summarizes the results of three experiments.
As can be seen from this table, our proposed radio SLAM is more accurate when compared to pure odometry.
For WiFi SLAM, indoor environments give better accuracy when compared to outdoor environment.
For example, in SUTD indoor and NTU semi-indoor, we obtain a localization accuracy of 3.69m and 4.48m, 
which are better than the accuracy of SUTD outdoor environment (8.82m).
Although a large number of APs are detected in outdoor environment, 
low signal strength values result in more uncertainty, 
since these APs inside buildings are far away from the moving path of the robot.
In general, WiFi SLAM gives better accuracy than LTE SLAM, 
due to the dense WiFi coverage in the environment. 
For NTU semi-indoor environment, we obtain an accuracy of 4.48m with WiFi SLAM, which is an improvement of 63.34\%
when compared to LTE SLAM (12.22m). 

\subsection{Evaluation of Radio+LiDAR SLAM}
\label{experiment_wifi_liar_slam}
Based on the optimized trajectory obtained from Radio SLAM, 
we perform LiDAR scan matching to further improve the accuracy based on the technique in Section \ref{radio_lidar_slam}. 
A valid match is identified if the fitness score is smaller than 0.1 and the matching points are larger than half of the average 
number of points in both scans. 
For NTU carpark semi-indoor environment,
our WiFi+LiDAR SLAM provided a localization accuracy of 0.84m,
while GMapping gave a localization accuracy of 4.31m and failed to correct odometry error.

We tested our approach on Intel Core i7-6700HQ CPU with 2.6GHz frequency and 8GB RAM. 
We observe that the time consumed for WiFi SLAM is insignificant as compared to time of data recording,
while the computational time for WiFi+LiDAR SLAM is much longer, due to the high cost of LiDAR scan matching. 
The time required for performing one WiFi similarity comparison with 44 MAC addresses is 0.029ms, while the time required for one LiDAR scan matching of 600 points
is 13.04ms. Therefore, LiDAR scan matching is the current bottleneck
of the proposed Radio+LiDAR SLAM. 

\begin{figure}
\centering
\includegraphics[width=0.498\textwidth]{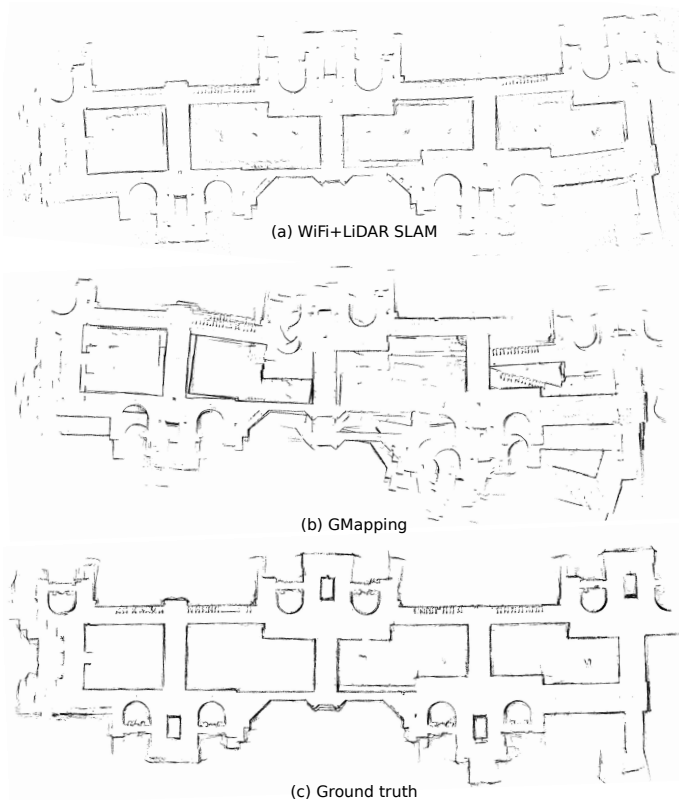}
\caption{
A comparison of the occupancy maps created by different approaches.
}
\label{fig_map}
\end{figure}

To visually inspect the quality of the optimized trajectory, 
we generate the occupancy grid map by evaluating the respective grid using LiDAR scans at optimized poses, as shown in Figure \ref{fig_map}. We also compared our results with GMapping, 
which applies particle filtering-based approach for SLAM. 
The map generated with GMapping is not consistent with the true map due to the drift of odometry. 
This can be seen from the non-overlapping boundaries representing the walls in Figure \ref{fig_map}(b). 
The quality of the 
map is significantly improved by the fusion of WiFi in Figure \ref{fig_map}(a). 

\begin{figure}
\centering
\includegraphics[width=0.498\textwidth]{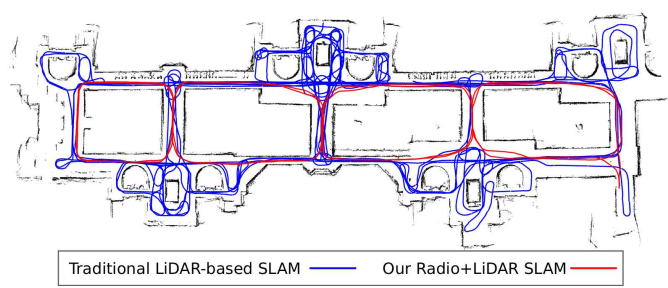}
\caption{Illustration of the path traveled by the conventional LiDAR-based SLAM (i.e., GMapping) and our proposed Radio+LiDAR SLAM for the mapping of a large indoor building.}
\label{path_different_approaches}
\end{figure}

A comparison of the path required for mapping large indoor building based on different approaches
is shown in Figure \ref{path_different_approaches}. 
To ensure the quality of map using the conventional LiDAR-based SLAM (i.e., GMapping), 
we have to design a cumbersome strategy for the robot (see the blue path in Figure\,\ref{path_different_approaches} with a total travel time of 8619.2s) to traverse different small regions and avoid large loops, 
which normally lead to mapping failure due to the lack of LiDAR loop closures to correct the large odometry drift after long distance travel. 
Our approach uses radio fingerprints to assist the conventional LiDAR-based SLAM. 
The proposed approach provides a comparable mapping quality when compared to the conventional LiDAR-based SLAM, but at a much faster scanning speed. 
It is important to point out that the deployment of radio infrastructure in the testing area is a necessity to perform the proposed radio SLAM.
The accuracy will be decreasing if the density of APs or base stations is low.

\section{Conclusions}
\label{conclusions}
This paper discussed the favorable sensors, popular algorithms, and critical challenges in SLAM. 
We proposed to use opportunistic WiFi and LTE signals, which are available in the existing infrastructure to perform SLAM. 
The performance of the system is verified in three different environments. 
We achieve a positioning accuracy of less than 10m using WiFi SLAM in all test environments. 
In addition, we present Radio+LiDAR SLAM that integrates LiDAR scan matching to improve the accuracy. 
The proposed approach produced comparable map quality versus LiDAR-based SLAM but at a much faster scanning speed. 
Our approach makes use of the existing radio infrastructure for sensing and has no assumption about the locations of the base stations.
In future, 
we would like to test our approach in different environments to validate the generalization of the proposed approach.
Another direction to investigate the possibility to use the created occupancy map for the navigation and path following 
of autonomous vehicles.

\section*{Acknowledgment}
This research is supported by  A*STAR under its RIE2020 Advanced Manufacturing and Engineering (AME) Industry Alignment Fund – Pre Positioning (IAF-PP) (Grant No. A19D6a0053). R. Liu's work is supported by the Natural Science Foundation of Sichuan Province under Grant 2023NSFSC0505. S. Mao's work is supported in part by the NSF under Grant ECCS-1923163. Any opinions, findings and conclusions or recommendations expressed in this material are those of the author(s) and do not reflect the views of A*STAR.


\begin{thebibliography}{}

\bibitem{graph_slam_tutorial}
G.~{Grisetti}, R.~{K\"ummerle}, C.~{Stachniss}, and W.~{Burgard}, ``A tutorial
  on graph-based slam,'' \emph{IEEE Intelligent Transportation Systems
  Magazine}, vol.~2, no.~4, pp. 31--43, 2010.

\bibitem{darpa}
M.~Tranzatto, T.~Miki, M.~Dharmadhikari, L.~Bernreiter, M.~Kulkarni,
  F.~Mascarich, O.~Andersson, S.~Khattak, M.~Hutter, R.~Siegwart, and
  K.~Alexis, ``Cerberus in the darpa subterranean challenge,'' \emph{Science
  Robotics}, vol.~7, no.~66, 2022.

\bibitem{suining_Chameleon}
S.~He, B.~Ji, and S.-H.~G. Chan, ``Chameleon: Survey-free updating of a
  fingerprint database for indoor localization,'' \emph{IEEE Pervasive
  Computing}, vol.~15, no.~4, pp. 66--75, 2016.

\bibitem{BearingSLAM}
A.~Arun, R.~Ayyalasomayajula, W.~Hunter, and D.~Bharadia, ``P2slam: Bearing
  based wifi slam for indoor robots,'' \emph{IEEE Robotics and Automation
  Letters}, vol.~7, no.~2, pp. 3326--3333, 2022.

\bibitem{wifi_crowdsensing}
Y.~Kim, Y.~Chon, and H.~Cha, ``Mobile crowdsensing framework for a large-scale
  wi-fi fingerprinting system,'' \emph{IEEE Pervasive Computing}, vol.~15,
  no.~3, pp. 58--67, 2016.

\bibitem{robust_fingerprint}
D.~Li, J.~Xu, Z.~Yang, C.~Wu, J.~Li, and N.~D. Lane, ``Wireless localization
  with spatial-temporal robust fingerprints,'' \emph{ACM Transactions on Sensor
  Networks}, vol.~18, no.~1, pp. 1--23, 2022.

\bibitem{ran_iot_2019}
R.~{Liu}, S.~H. {Marakkalage}, M.~{Padmal}, T.~{Shaganan}, C.~{Yuen}, Y.~L.
  {Guan}, and U.~{Tan}, ``Collaborative slam based on wifi fingerprint
  similarity and motion information,'' \emph{IEEE Internet of Things Journal},
  vol.~7, no.~3, pp. 1826--1840, March 2020.

\bibitem{OrbSLAM3}
C.~Campos, R.~Elvira, J.~J.~G. Rodr\'iguez, J.~M. M.~Montiel, and
  J.~D.~Tard\'os, ``Orb-slam3: An accurate open-source library for visual,
  visual–inertial, and multimap slam,'' \emph{IEEE Transactions on Robotics},
  vol.~37, no.~6, pp. 1874--1890, 2021.

\bibitem{superPoint}
D.~DeTone, T.~Malisiewicz, and A.~Rabinovich, ``Superpoint: Self-supervised
  interest point detection and description,'' in \emph{2018 IEEE/CVF Conference
  on Computer Vision and Pattern Recognition Workshops}, June 18--22 2018, pp.
  337--33\,712.

\bibitem{mmwave_tutorial}
A.~Shastri, N.~Valecha, E.~Bashirov, H.~Tataria, M.~Lentmaier, F.~Tufvesson,
  M.~Rossi, and P.~Casari, ``A review of millimeter wave device-based
  localization and device-free sensing technologies and applications,''
  \emph{IEEE Communications Surveys \& Tutorials}, vol.~24, no.~3, pp.
  1708--1749, 2022.

\bibitem{mmwave_mapping}
C.~Baquero~Barneto, E.~Rastorgueva-Foi, M.~F. Keskin, T.~Riihonen, M.~Turunen,
  J.~Talvitie, H.~Wymeersch, and M.~Valkama, ``Millimeter-wave mobile sensing
  and environment mapping: Models, algorithms and validation,'' \emph{IEEE
  Transactions on Vehicular Technology}, vol.~71, no.~4, pp. 3900--3916, 2022.

\bibitem{distributed_mapping_wifi}
C.~{Adhivarahan} and K.~{Dantu}, ``{WISDOM: WIreless Sensing-assisted
  Distributed Online Mapping},'' in \emph{2019 International Conference on
  Robotics and Automation}, May 20--24 2019, pp. 8026--8033.

\bibitem{Ismail_case2022}
K.~Ismail, R.~Liu, Q.~Zhenghong, A.~Athukorala, B.~P.~L. Lau, M.~Shalihan,
  C.~Yuen, and U.-X. Tan, in \emph{Efficient WiFi LiDAR SLAM for Autonomous
  Robots in Large Environments}, August 22--26 2022, pp. 1132--1137.

\bibitem{KurzPrunningIROS2021}
G.~Kurz, M.~Holoch, and P.~Biber, ``Geometry-based graph pruning for lifelong
  slam,'' in \emph{2021 IEEE/RSJ International Conference on Intelligent Robots
  and Systems (IROS)}, 2021, pp. 3313--3320.

\bibitem{Mixture_model}
T.~{Pfeifer} and P.~{Protzel}, ``Expectation-maximization for adaptive mixture
  models in graph optimization,'' in \emph{2019 International Conference on
  Robotics and Automation}, May 20-24, pp. 3151--3157.

\bibitem{joint_compatibility}
J.~{Neira} and J.~D. {Tardos}, ``Data association in stochastic mapping using
  the joint compatibility test,'' \emph{IEEE Transactions on Robotics and
  Automation}, vol.~17, no.~6, pp. 890--897, 2001.

\bibitem{Pairwise_outlier_rm}
J.~G. {Mangelson}, D.~{Dominic}, R.~M. {Eustice}, and R.~{Vasudevan},
  ``Pairwise consistent measurement set maximization for robust multi-robot map
  merging,'' in \emph{2018 IEEE International Conference on Robotics and
  Automation}, May 21--25 2018, pp. 2916--2923.

\bibitem{Semantics_mapping}
S.~Garg, N.~S\"underhauf, F.~Dayoub, D.~Morrison, A.~Cosgun, G.~Carneiro,
  Q.~Wu, T.-J. Chin, I.~Reid, S.~Gould, P.~Corke, and M.~Milford, ``Semantics
  for robotic mapping, perception and interaction: A survey,''
  \emph{Foundations and Trends® in Robotics}, vol.~8, no. 1–2, pp. 1--224,
  2020.
\end{thebibliography}

\begin{IEEEbiographynophoto}{Ran Liu} received the Ph.D. from University of T\"uebingen, Germany, in 2014. He is an associate professor at Southwest University of Science and Technology, China. His research interests include robotics and SLAM. Contact him at ran.liu.86@hotmail.com
\end{IEEEbiographynophoto}
\begin{IEEEbiographynophoto}{Billy Pik Lik Lau} received the Ph.D. from Singapore University of Technology and Design in 2021. He is working as a Research Fellow at the Singapore University of Technology and Design. His research focus includes smart city and internet of things. Contact him at billy\_lau@sutd.edu.sg
\end{IEEEbiographynophoto}

\begin{IEEEbiographynophoto}{Khairuldanial Ismail} is currently pursuing a Ph.D. degree with Singapore University of Technology and Design. His research interests include robotics and indoor localization. Contact him at khairuldanial\_ismail@sutd.edu.sg
\end{IEEEbiographynophoto}

\begin{IEEEbiographynophoto}{Achala Chathuranga} received his bachelor's degree from University of Moratuwa, Sri Lanka in 2018. He is currently working as a Research Engineer at Singapore University of Technology and Design. His research interests are mobile robot localization and path planning. Contact him at achala\_chathuranga@sutd.edu.sg
\end{IEEEbiographynophoto}

\begin{IEEEbiographynophoto}{Chau Yuen} (\textit{Corresponing author}) received his Ph.D. from Nanyang Technological University, Singapore in 2004. From 2010 to 2023, he was with the Engineering Product Development Pillar, Singapore University of Technology and Design. Since 2023, he has been with the School of Electrical and Electronic Engineering, Nanyang Technological University. He serves as an Editor-in-Chief for Springer Nature Computer Science, Editor for IEEE TRANSACTIONS ON VEHICULAR TECHNOLOGY, IEEE SYSTEM JOURNAL, and IEEE TRANSACTIONS ON NETWORK SCIENCE AND ENGINEERING. Contact him at chau.yuen@ntu.edu.sg
\end{IEEEbiographynophoto}

\begin{IEEEbiographynophoto}{Simon X. Yang} received his Ph.D. from the University of Alberta, Edmonton, Canada, in 1999. He is currently a professor at University of Guelph, Canada. He is currently the Editor-in-Chief for the International Journal of Robotics and Automation and an Associate Editor for the IEEE Transactions on Cybernetics. Contact him at syang@uoguelph.ca
\end{IEEEbiographynophoto}

\begin{IEEEbiographynophoto}{Yong Liang Guan} obtained his Ph.D. from the Imperial College London, UK. He is a professor of Communication Engineering at the School of Electrical and Electronic Engineering, Nanyang Technological University, where he leads the Continental-NTU Corporate Research Lab. He is an Editor for the IEEE Transactions on Vehicular Technology. Contact him at eylguan@ntu.edu.sg
\end{IEEEbiographynophoto}

\begin{IEEEbiographynophoto}{Shiwen Mao} received his Ph.D. from Polytechnic University, Brooklyn, New York, in 2004. He is a professor and Earle C. Williams Eminent Scholar, and Director of the Wireless Engineering Research and Education Center at Auburn University.
He is the Editor-in-Chief of IEEE Transactions on Cognitive Communications and Networking.
Contact him at smao@ieee.org
\end{IEEEbiographynophoto}

\begin{IEEEbiographynophoto}{U-Xuan Tan} received his Ph.D. degrees from Nanyang Technological University in 2010. 
He is an associate professor at the Singapore University of Technology and Design. 
He serves as an associate editor for IEEE Robotics and Automation Letters. Contact him at uxuan\_tan@sutd.edu.sg
\end{IEEEbiographynophoto}

\end{document}